# Pan-infection Foundation Framework Enables Multiple Pathogen Prediction

Lingrui Zhang[1]#, Haonan Wu[2]#, Nana Jin[2], Chenqing Zheng[2], Jize Xie[3], Qitai Cai[2], Jun Wang[4], Qin Cao[5], Xubin Zheng[6]*, Jiankun Wang[1,7]*, Lixin Cheng[2]*

[1] Shenzhen Key Laboratory of Robotics Perception and Intelligence, and the Department of Electronic and Electrical Engineering, Southern University of Science and Technology, Shenzhen, China

[2] Department of Critical Care Medicine, Shenzhen People's Hospital, The First Affiliated Hospital of Southern University of Science and Technology, Shenzhen 518020, China

[3] Department of Industrial Engineering and Decision Analytics, The Hong Kong University of Science and Technology, Hong Kong

[4] Bioinformatics Centre, Department of Biology, University of Copenhagen, København Ø 2100, Denmark

[5] School of Biomedical Science, The Chinese University of Hong Kong, Hong Kong

[6] Great Bay Area University, Guangdong, China

[7] Jiaxing Research Institute, Southern University of Science and Technology, Jiaxing, China.

**Abstract**
Host-response-based diagnostics can improve the accuracy of diagnosing bacterial and viral infections, thereby reducing inappropriate antibiotic prescriptions. However, the existing cohorts with limited sample size and coarse infections types are unable to support the exploration of an accurate and generalizable diagnostic model. Here, we curate the largest infection host-response transcriptome data, including 11,247 samples across 89 blood transcriptome datasets from 13 countries and 21 platforms. We build a diagnostic model for pathogen prediction starting from a pan-infection model as foundation (AUC = 0.97) based on the pan-infection dataset. Then, we utilize knowledge distillation to efficiently transfer the insights from this "teacher" model to four lightweight pathogen "student" models, *i.e.*, staphylococcal infection (AUC = 0.99), streptococcal infection (AUC = 0.94), HIV infection (AUC = 0.93), and RSV infection (AUC = 0.94), as well as a sepsis "student" model (AUC = 0.99). The proposed knowledge distillation framework not only facilitates the diagnosis of pathogens using pan-infection data, but also enables an across-disease study from pan-infection to sepsis. Moreover, the framework enables high-degree lightweight design of diagnostic models, which is expected to be adaptively deployed in clinical settings.

Infections are well-documented causes of numerous diseases, ranging from mild to severe. For instance, staphylococcus aureus bacteremia (SaB) caused by staphylococcal infections, presents a substantial challenge to healthcare systems, with mortality rates as high as 20% to 30%[1][2][3]. Critically, the incidence of SaB infections is increasing in recent years, which further exacerbates its severity as a public health threat[4]. Streptococcal infections also pose serious risks, as these can infect various host organisms and tissues and are transmissible between humans and animals[5]. Most *streptococci* cause problems in the functioning of the human respiratory or gastrointestinal tract and cause some of the most critical infections, such as sepsis[6]. Sepsis is a life-threatening reaction to a variety of infections that causes inflammation and organ damage (Fig.1a), kills about 270,000 people each year in the USA[7]. Timely and accurate prediction and treatment of infection and disease are critical for patients[8].

Infection induces measurable changes in gene expression within the host, providing valuable insights into the immune response to various pathogens[9][10]. By analyzing gene expression profiles in blood samples, researchers can implement targeted therapeutic interventions based on the host's response[11][12][13][14]. Recently, a number of approaches have emerged that utilize this gene expression profile for the prediction of infections, such as the prognostic models proposed by[15][16][17][18][19]. Network-based approaches and moonlighting long noncoding RNAs (lncRNAs) have also been used for prediction[20]. However, these existing models face two significant limitations: 1) the small size of training datasets, which hinders their generalizability and increases the risk of overfitting; 2) the coarse classification of infections, typically distinguishing only between viral and bacterial infections without specific types of pathogens involved. This lack of specificity can lead to inappropriate treatment decisions, as different pathogens may require distinct therapeutic approaches, resulting in the

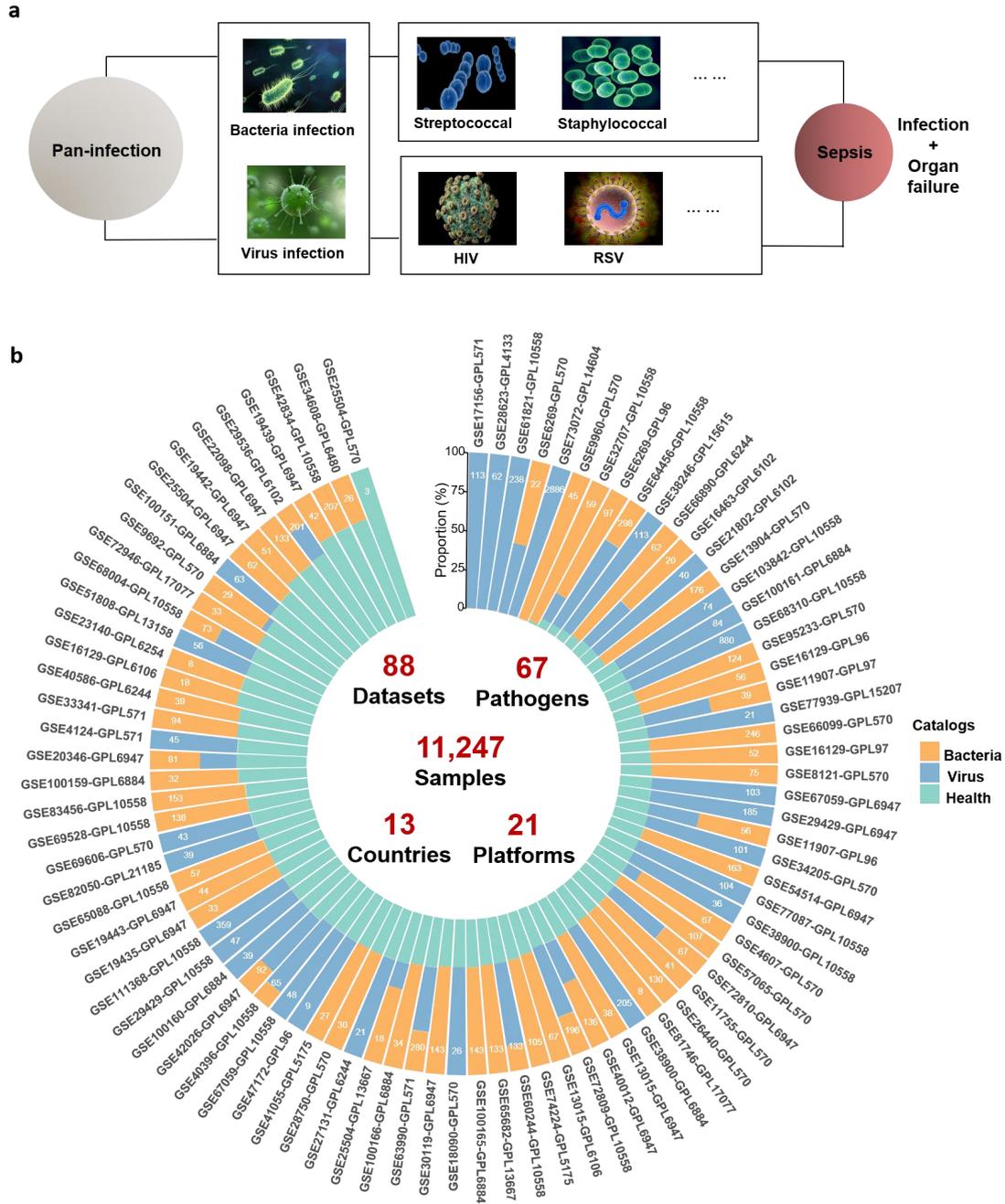

**Fig.1 | Data used in this study. a,** overview of pan-infection and sepsis. **b,** 88 host transcriptome datasets from 13 countries and 21 platforms containing 11,247 samples across 67 pathogens and sepsis. Datasets are sorted by the ratio of health samples.

misuse of antibiotics[21]. Fortunately, the increasing availability of larger datasets and advancements in Artificial Intelligence (AI) in recent years present a great opportunity to apply more sophisticated analytical methods on expansive datasets, paving the way for the development of more precise and robust diagnostic models for different types of infections.[22][23]

Knowledge Distillation (KD) is a sophisticated AI method that performs the lightweight model (known as the student) to learn informative knowledge from the well-trained yet cumbersome model (known as the teacher)[24][25][26][27]. This approach possesses the potential for effective infection diagnosis, as it allows models to learn from diverse infections, enhancing their accuracy even with limited data. Additionally, the reduced complexity of the student model enables its deployment in real-world clinical applications with limited computational resources, facilitating timely and accurate diagnostics.

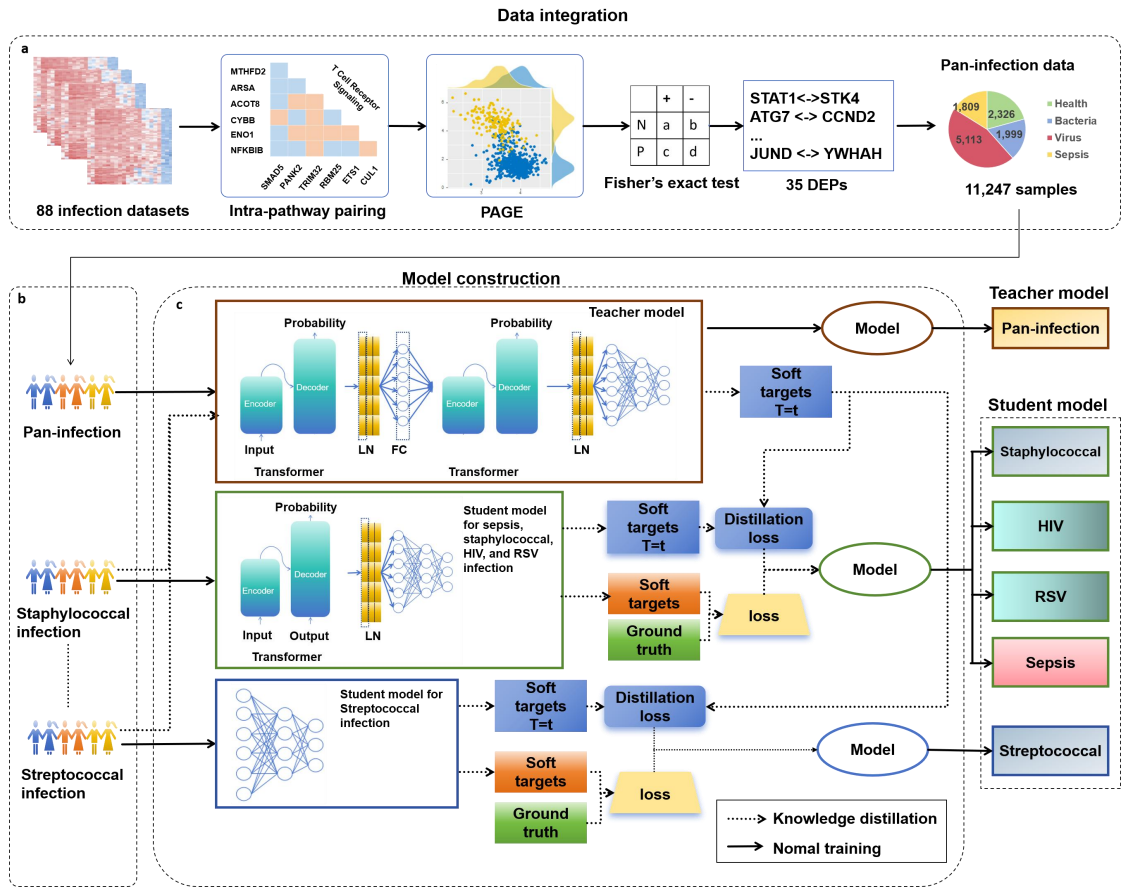

**Fig.2 | Workflow of TSGPS. a**, Construction of pan-infection data. 11,247 transcriptome samples were integrated using PAGE with intra-pathway pairing and Fisher's exact test. 35 DGPs were screened out for subsequent analysis. **b**, Samples were divided into different sets according to infection types. **c**, Three different models were designed through the transformer layer, layer norm (LN), fully connected layer (FC), and multi-layer perceptron (MLP). The student models were trained by combining distillation loss and ground truth loss.

In this study, we harness KD, and pan-infection data for multiple pathogen prediction and infection-related disease diagnosis. After curating an integrated pan-infection dataset of over ten thousand samples, we used a coarse-to-fine teacher-student architecture based on KD and defined it as Teacher-Student Gene Pair Signature (TSGPS), which reduces the number of parameters of the model while improving prediction accuracy and facilitating hospital deployment. TSGPS can simultaneously and precisely diagnose six diseases, including pan-infection, staphylococcal infection, streptococcal infection, HIV, RSV, and sepsis.

## Results

### Establishment of pan-infection data.

To construct a most comprehensive host transcriptome dataset, we collected 88 distinct datasets from the Gene Expression Omnibus (GEO) database[5], which collectively encompass 11,247 samples from 13 countries and 21 diverse platforms. These samples are categorized into three primary groups, 1,505 bacterial infection samples, 5,113 viral infection samples, 1,809 sepsis samples, and 2,326 healthy controls (Fig.1b).

The datasets encompass a diverse array of diseases stemming from 67 distinct pathogens, including bacteria, such as staphylococcal infection, streptococcal infection, HIV infection, and RSV infection, as well as viruses, such as HIV and RSV. We integrated and processed the collected data using intra-pathway pairing and PAGE[7]. We selected 35 differential gene pairs (DGPs) for the pan-infection dataset and then selected five DGP sets with the same size corresponding to sepsis, staphylococcal infection, streptococcal infection, HIV, and RSV respectively. This comprehensive dataset enables training of model from a wide range of pathological conditions, thereby improving its ability to generalize and accurately classify novel infections.

**The structure of TSGPS.**

We proposed a novel method, Teacher-Student Gene Pair Signature (TSGPS), to parallelly execute multiple tasks for pathogen identification and disease diagnosis using the teacher-student architecture. The workflow of TSGPS consists of two sections, one pan-infection "teacher" model construction and multiple pathogen "student" model constructions (Fig.2). Firstly, we used pan-infection data to train a pan-infection "teacher" model, which constructed based on transformer[28] modules with a proven ability in modeling global feature. Equipped with this foundation teacher model, we constructed two different simplified student models based on lightweight transformer and MLP, respectively. Then, the well-trained pan-infection foundation model was performed in distilling specific student models for four pathogen infections, as well as for cross-disease prediction. During the training, we designed a distillation loss to transfer knowledge from the foundation teacher model to each pathogen student models. This process enables feature transfer between various infections, making TSGPS highly scalable. Finally, our proposed TSGPS distills five highly specialized lightweight student models, including staphylococcal infection, streptococcal infection, HIV infection, RSV infection, and sepsis, which can be easily deployed in clinical scenarios for infection diagnosis.

**Pan-infection foundation model.**

Integrating diverse infection data for pan-infection infrastructures faces challenges due to limited training data, impacting classifier accuracy[29][30]. Biologically, infections trigger

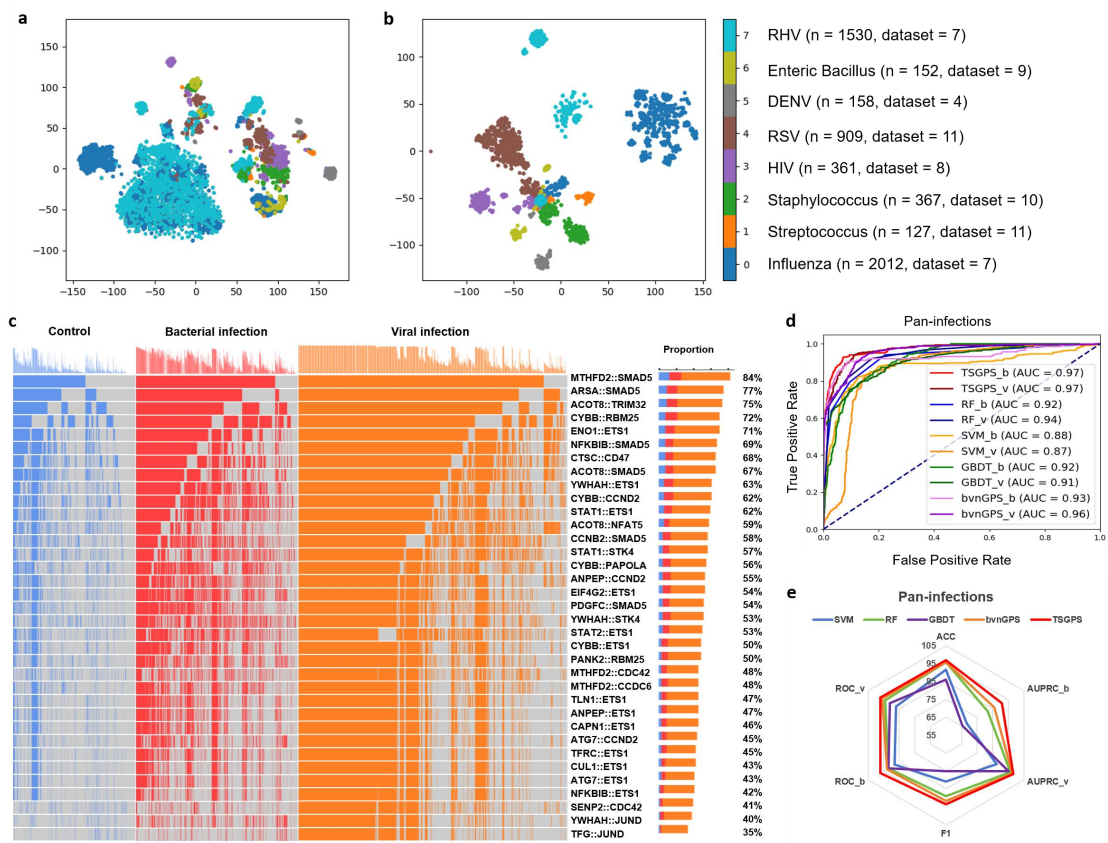

**Fig.3 | Characterization and performance of the pan-infection foundation model. a, b**, the visualization of samples before (a) and after (b) screening out gene pairs with PAGE. Eight pathogens with the largest sample size are shown. Samples of the same pathogen tend to cluster together after PAGE. **c,** Waterfall plot showing the expression pattern of the 35 gene pairs across phenotypes in the pan-infection dataset. **d,** ROC comparison with other models. TSGPS_b: teacher model for bacterial infection; TSGPS_v: the teacher model for virus infection. **e,** Performance comparison using ACC, F1 score, AUC, AUPRC, and precision. ROC_b and AUPRC_b for bacterial infections. ROC_v and AUPRC_v for virus infections.

**Table 1 | Data used for model construction**

| Disease | Health | Infection | Total |
|---|---|---|---|
| Pan-infection | 985 | Bacteria: 1,505<br><br>Virus: 3,843 | 6,333 |
| Sepsis | 448 | 1,809 | 2,257 |
| streptococcal infection | 179 | 127 | 306 |
| staphylococcal infection | 318 | 367 | 685 |
| HIV | 230 | 361 | 591 |
| RSV | 166 | 909 | 1,075 |

The details of pan-infection data, sepsis data, streptococcal infection data, staphylococcal infection data, HIV data, and RSV data.

immune responses, altering RNA. A foundation model aids early bacterial/viral diagnosis and supports the training of specific infection models. Given the sensitive host response to the infection caused by different pathogens, several studies concentrated on analyses best-informed by host transcriptome data. To this end, we integrated 11,247 samples using PAGE and identified 35 DGPs in infection. Samples of the same infection type tend to cluster together after PAGE, while these samples were scattered before PAGE (Fig.3a-b), demonstrating the effectiveness of PAGE in data integration across platforms. The curated pan-infection data was used to construct six infection-related models (Table 1).

Teacher-student architecture was used to build six infection-related models, where the pan-infection foundation model (PIFM) served as the "teacher" and pathogen models served as "students". The PIFM is a crucial determinant for the subsequent performance of the other student models. We trained the "teacher" model with 80% of the pan-infection samples. The PIFM applied a deep learning neural network with self-attention mechanisms, layer normalization, fully connected layers, and MLP.

To rigorously assess the performance of the PIFM, we reserved 20% of the pan-infection samples as validation set to evaluate its performance using five metrics, including Accuracy (ACC), Area Under Curve (AUC), F1 Score, Precision, and Area Under the Precision-Recall Curve (AUPRC). PIFM demonstrates superior performance, achieving an AUC of 0.97 for both bacterial and viral infections (Fig.3d & 3e). We compared PIFM with bacterial-viral-noninfected GPS (bvnGPS)[14], Random Forest (RF), Gradient Boosting Decision Tree (GBDT), and Support Vector Machine (SVM). PIFM surpassed RF by at least 0.03 and SVM by at least 0.09 in AUC. Overall, it outperforms these methods across all evaluation metrics, highlighting the effectiveness of TSGPS (Fig.3e).

**Pan-infection foundation model enhances pathogen prediction.**

After building pan-infection foundation model, we transferred this "teacher" model to establish simple and small pathogen student models for specific pathogen prediction. The student models applied fewer transformer layers and network layers. We conducted a series of rigorous experiments utilizing specific infection datasets to demonstrate the capability of our proposed framework to enhance the performance of diverse models.

We concentrated on four common infections, *i.e.*, staphylococcal infection, streptococcal infection, HIV, and RSV. TSGPS was separately used for training the corresponding prediction model. For comparison, we also train four vanilla models without using TSGPS. The model with the assistance of the teacher model consistently showed higher AUC scores compared to the vanilla model. The AUCs of TSGPS in the Staphylococcus infection model, Streptococcus infection model, HIV model, and RSV model were 0.99, 0.94, 0.93, and 0.94, respectively, while the AUCs of the vanilla model were 0.95, 0.93, 0.88, and 0.89, respectively. Additionally, we also trained RF, GBDT, and SVM models for comparison, the AUCs of which were 0.93 to 0.95, 0.86 to 0.91, 0.84 to 0.92, and 0.86 to 0.92 for staphylococcal infection, streptococcal infection, HIV, and RSV respectively. (Fig.5c).

The comprehensive evaluation of our model and structure encompasses various indicators, providing a multi-faceted assessment of their performance (Fig.5d). Our result shows the performance of the vanilla model improves significantly when harnessed TSGPS, even the vanilla model is not the optimal one across all metrics. For instance, in models trained with staphylococcal infection data, the Random Forest model outperforms our original model in terms of the AUPRC, achieving a score of 0.96 compared to our original model's 0.94. However, the student model, learning from the PIFM, outperforms all the infection models, achieving an AUPRC score of 0.99. A similar trend is observed in the HIV data, where TSGPS outperforms other models across all evaluation metrics. Specifically, TSGPS consistently enhances the performance of the original student model, leading to improvements in ACC, F1 Score, precision, AUPRC, and AUC. These findings underscore the effectiveness of our structure in enhancing the capabilities of various modeling paradigms, thereby demonstrating its potential to yield superior results across the analysis of different infections.

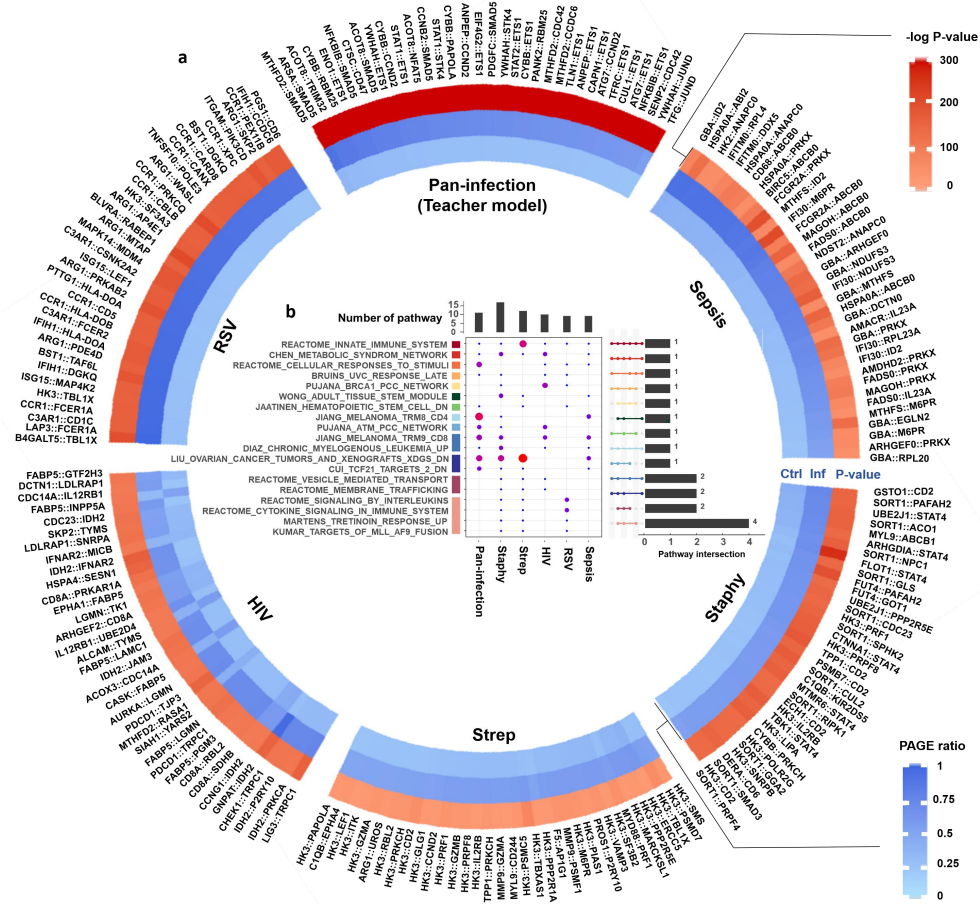

**Fig.4 | Model composition of the six models.** Gene pairs associated with the six infections are summarized. **a**, Heatmap showing the PAGE ratio of patient samples, the PAGE ratio of healthy samples, and corresponding log *p*-value from inside to outside. **b**, Pathway analysis and upset plot showing the functional association among different infections. The PAGE ratio is defined as the ratio of samples with specific pairwise expression pattern (G1 > G2) in each phenotype of a given cohort.

**Pan-infection foundation model enhances sepsis diagnosis.**

Besides pathogen prediction, the pan-infection foundation can also be applied to enhance disease diagnosis such as sepsis. Given the high mortality of sepsis and its potential to be triggered by bacterial or viral infections, early screening is crucial for timely treatment. Host immune system has an extreme response during the development from infection to sepsis, indicating host immune response is a common knowledge across the two diseases. Similarly, the expression pattern of DGPs shares common knowledge between teacher and student models (Fig.6e). In this work, transmission of cross-disease features from teacher model to student model corresponds to the transition from pan-infection to sepsis.

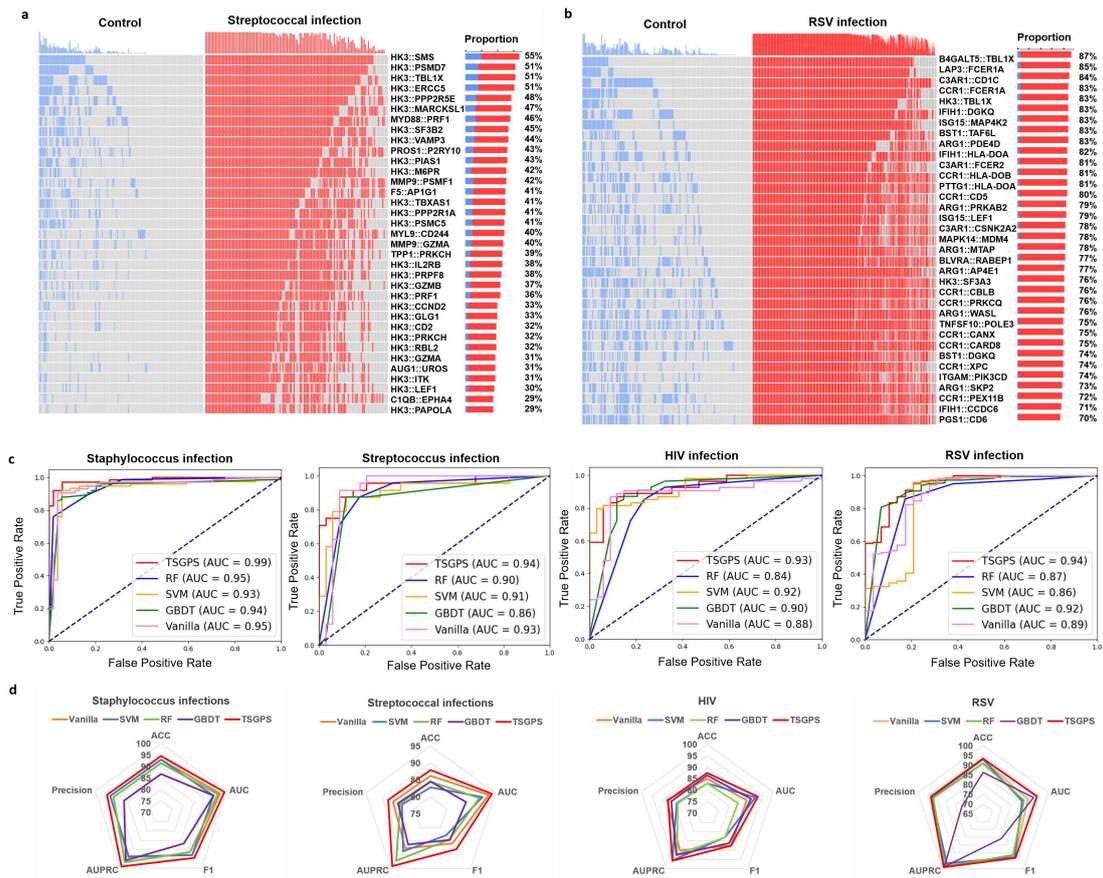

**Fig.5 | Performance in pathogen infection prediction. a**, **b**, Waterfall plots showing the expression pattern of the 35 gene pairs across phenotypes in the streptococcal infection and RSV datasets. **c**, Comparison of ROC curves with other models. **d**, Performance comparison using ACC, F1 score, AUC, AUPRC, and precision.

The constructed sepsis student model contained fewer attention layers and heads with distillation loss and cross-entropy loss (see Methods). Validation on independent datasets demonstrated the performance of the sepsis student model enhanced by the pan-infection model (Fig.6b-c). Compared to the original model without KD, the AUC of the student model improves by 0.02 and the ACC metric improves by 0.03 (reaching 0.99 and 0.96, respectively). Furthermore, we compared it to the existing biomarkers, SeptiCyte[16] and sNIP[31], which are rapid molecular detection methods that can differentiate between healthy individuals and those with sepsis. The performance on four-fold cross validation was evaluated using AUPRC, ACC, and AUC, where TSGPS demonstrates superior robustness and consistently achieves higher AUPRC, ACC, and AUC scores compared to SeptiCyte and sNIP (Fig.6d). The effectiveness of TSGPS for cross-disease demonstrates the potential for early diagnosis of disease.

**Characterization of TSGPS.**
With the assistance of TSGPS, the five "student" models demonstrated superior performance utilizing 35 DGPs. We investigated the DGPs we screened across pan-infection, staphylococcal infection, streptococcal infection, HIV, RSV, and sepsis (Fig.7a). The DGPs in different circumstances have few intersects, which explains the high efficiency of the model as they may focus on different aspects. The DGPs we screened for pan-infection and various other infections share the same gene ontology (GO) terms at the functional level (Fig.4), which shows the plausibility of TSGPS in terms of knowledge transfer. The number of these DGPs is much higher in the infected population than in the healthy population, the probability of the abnormal DGPs being present in the population with the disease is essentially 70% or more (Fig.3c, Fig.5a, Fig.5b). All these results explain our model from a bioinformatics point of view.

To explore the effectiveness of KD in pathogen and disease prediction, we analyzed the pathways and functions of the KD "teacher"

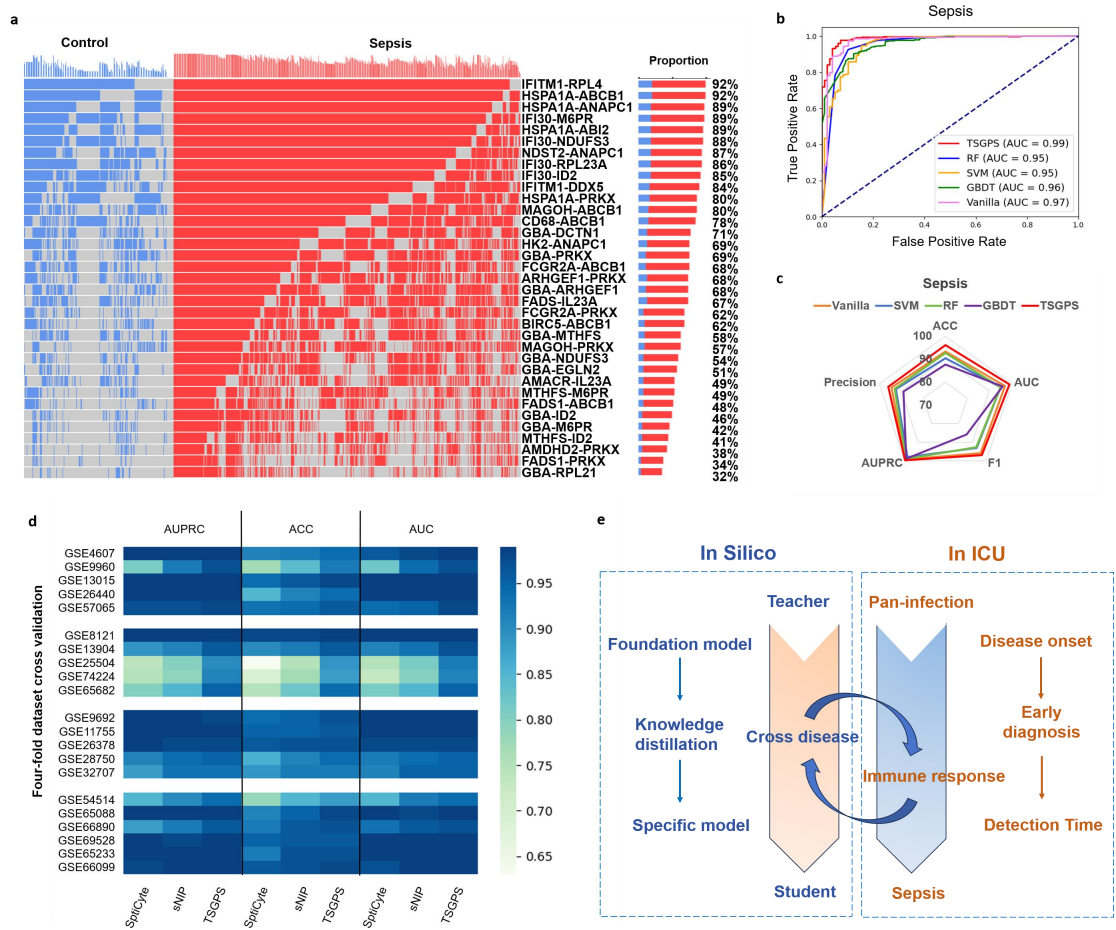

**Fig. 6 | The performance of sepsis student model. a,** Waterfall plot showing the expression pattern of the 35 gene pairs across phenotypes in the sepsis dataset. **b,** ROCs comparison with other models. **c,** Performance comparison using ACC, F1 score, AUC, AUPRC, and precision. **d,** Performance comparison with SOTA, including TSGPS, SeptiCite, and sNIP, using four-fold dataset cross validation. **e,** Rationale between infection and sepsis as well as principle of KD.

and "student" models. Streptococcal and staphylococcal infections, which are both classified as bacterial infections, share 10 common genes. Beyond this, only a limited number of genes were identified as overlapping across the various diseases (see Fig. 7a). However, the pathways enriched by these genes were found to be consistent across the various diseases. Pan-infection shares 24 pathways with other infections or diseases, mainly focused on the production of inflammatory cytokines, which are crucial for immune response. Notable pathways include the JAK-STAT signaling pathway, the WNT signaling pathway, and the T cell receptor signaling pathway. Infections caused by different pathogens share several pathways, particularly those related to innate immunity and inflammation, such as leukocyte transendothelial migration. Additionally, pathways associated with cell adhesion, including tight junction and cell adhesion molecules (CAMs), are also commonly shared (Fig. 7f).

Each infection is characterized by distinct DGPs that serve specific functions. For instance, DGPs associated with HIV are enriched in pathways that regulate cellular behavior and communication, including extracellular matrix organization, syndecan interactions, and the E2F pathway. In the case of streptococcal infections, DGPs are notably enriched in pathways that regulate cell differentiation, proliferation, and immune responses. Key pathways include CTLA4 inhibitory signaling, MAPK family signaling cascades, and NOTCH signaling. Staphylococcal infections exhibit DGPs enriched in immune responses and inflammation, such as the STAT4 pathway, TLR4 signaling and tolerance mechanisms, as well as interleukin 37 signaling. Additionally, DGPs in RSV are enriched in pathways related to immune responses and cellular activation, including the

CD40 pathway, T cell receptor signaling, and negative regulators of DDX5-IFIH1 signaling.

Additionally, higher semantic similarity scores indicate that diseases share related biological functions or pathways, while a greater number of shared connections implies stronger interactions within the PPI network (Fig.7b-e). The semantic similarity score and the number of connections shared between each disease in a PPI network suggest a potential transfer of features among diseases.

## Discussion

This study implements a pan-infection pretrained foundation and fine-tune framework for predicting multiple pathogens. We proposed and validated the effectiveness of the teacher-student model architecture in categorizing diverse infections by training five independent student models of distinct pathogens and diseases. This approach yields impressive performance with AUCs of 0.99, 0.94, 0.93, 0.94, and 0.99 for staphylococcal infections, streptococcal infections, HIV, RSV, and sepsis, respectively. Clinicians can utilize the predictions from the pan-infection model to optimize the allocation of medical resources, provide accurate early-stage treatments, minimize the resources waste and ultimately reduce patient mortality.

The KD architecture adopted by TSGPS facilitates knowledge transfer between different models. The teacher-student architecture we employed offers several advantages. First, it leverages a high-accuracy foundational teacher model with strong generalization capabilities to effectively guide the training of the student model. This allows for the transfer of not only output distributions but also intermediate, relational, and structural features, resulting in a more comprehensive knowledge transfer. Second, the student models consume fewer resources for storage and inference, which is important for deploying deep learning models in resource-constrained environments.

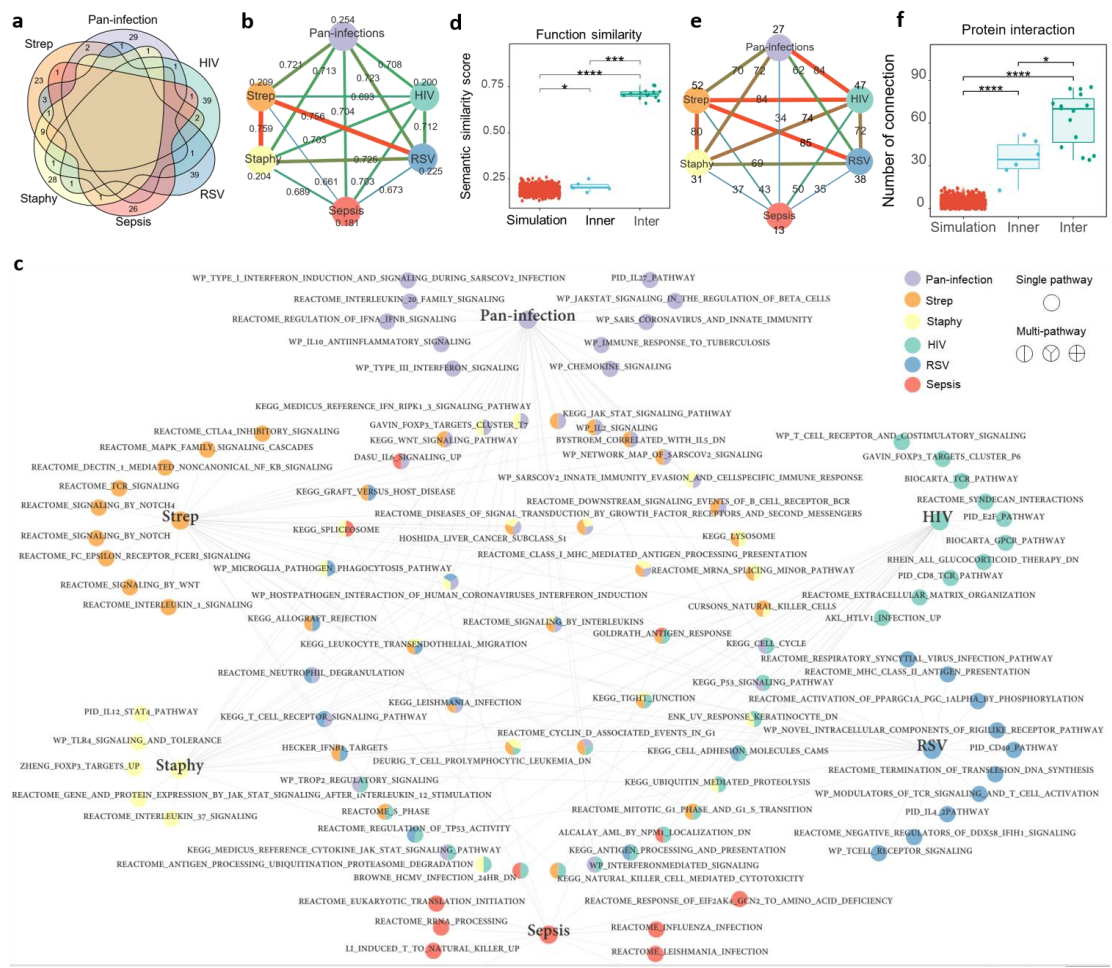

**Fig. 7 | Functional characterization of the genes of the six models. a,** The Venn diagram of the gene pair across the six infections and diseases. None of these six different infections and diseases share common gene pairs at the genetic level. **b, d,** The relationship between these six diseases in semantic similarity and protein interaction. **e, f,** The statistics of these six diseases performed on semantic similarity and protein interaction. A strong correlation can be observed between these six infections and diseases. **c,** The relationship of the genes involved in the pathways in each disease. These infections and diseases share many common pathways, with up to four infections sharing the same pathway.

We curated a pan-infection dataset comprising 11,247 samples from 88 sources, covering 13 countries and 21 platforms. The samples of the established pan-infection data come from various sources, presenting significant challenges in extracting cross-platform information using conventional methodologies. To address this problem, we utilize PAGE to integrate these cross-platform data sets, which is a sophisticated method we previously proposed for feature selection and data fusion[7][32][33]. With-in sample comparison of genes enables a more comparable quantification analysis for datasets from various platforms.

As the immune system has similar response to different pathogen infections, training the pan-infection teacher model can benefit the diagnosis of the specific pathogens. Besides, sepsis is a disease caused by infection therefore the pan-infection teacher model can help the student model in diagnosing sepsis. The integrated large-scale pan-infection dataset facilitates the training of deeper neural networks, specifically teacher-student architecture in this study, which can assist in parallelly and accurately detecting multiple diseases. Our results demonstrate the substantial potential of our framework in accurately detecting a broad spectrum of diseases. This framework is also characterized by ease of deployment. However, the scarcity of certain pathogen data has prevented TSGPS from providing accurate predictions for some cohorts with extremely small sample size. We anticipate that this limitation will improve as more data becomes available in the future.

We found that simpler models outperform complex models when trained on small datasets. For instance, in the streptococcal infection model, the one using the self-attention mechanism achieved an AUC of 0.89, while the one without self-attention mechanism achieved an AUC of 0.93.

Although TSGPS has modeled four pathogens and one infection disease, a wider range of pathogen screening and infection diagnosis are critical for further investigation, which calls for a more powerful pan-infection foundation model using a larger volume of data.

On the other hand, although TSGPS has achieved an AUC of 0.94 in cases of streptococcal infections, the accuracy of models can be further improved through other technical methods, such as incorporating few-shot learning methods. In this case, TSGPS can be used in some pathogens with fewer data. Moreover, the pathogen student models in TSGPS operate independently while sharing a common pan-infection model for knowledge acquisition. To further extend the application of this framework, we are planning to propose a novel scenario focused on differentiating between various bacterial and viral infections in the near future.

While the results obtained from TSGPS are promising, such as achieving an AUC of 0.94 for streptococcal infections, there is potential for further improving model accuracy through additional technical methods, including the incorporation of few-shot learning techniques. This approach would enable TSGPS to be effectively applied to pathogens with limited data. Meanwhile, a wider range of pathogen screening and infection diagnosis are critical for further investigation, which calls for a more powerful pan-infection foundation model using a larger volume of data.

Overall, we proposed a novel framework TSGPS, which has demonstrated strong performance in four cross-infection and one cross-disease studies, with limited number of parameters in the student models to ensure the deployment in clinical settings. The experiments exemplify the potential of our framework to accurately identify pathogens and predict diseases, underscoring its promising applications in clinical practice.

## Methods
### Study design
KD is a technique used in machine learning, particularly within the domain of neural networks and deep learning, where the knowledge from a larger model (often referred to as the "teacher" or "mentor" model) is transferred to a smaller model (often called the "student" model). The primary aim of this process is to enable the student model to mimic the performance and behavior of the teacher model while being more compact, faster, and less resource-intensive.

Existing distillation methods can be categorized into online distillation[34][35], self-distillation[36][37][38] and offline distillation[24][25][39][40]. The student model was designed in four ways: a quantitative version of the teacher model that preserves the network

structure[41][42], a simplified teacher model[27][43], a simple model with good performance[44][45] and an optimized global network[46][47]. However, the substantial amount of data required for these techniques limits its widespread application in bioinformatics. Building on the offline destillation, we propose a framework aimed at obtaining fine-grained and easy-to-deploy infection models. This framework revolves around a comprehensive teacher model that possesses the capability to discern between three fundamental categories: health, bacterial infection, and virus infection. The teacher model transferred the feature information to multiple student models, allowing them to achieve higher accuracy in their respective disease prediction tasks (Fig.2).

**TSGPS**

**PAGE and Data processing.** The collected cohorts can be divided into six categories, including, pan-infections, sepsis, streptococcal infection, staphylococcal infection, HIV, and RSV (Table 1). To integrate samples from different cohorts and platforms, we employed the PAGE algorithm[7] to identify differences between gene pairs. Each pair was constructed within the pathways in the Molecular Signatures Database (MSigDB). To build foundation model we assigned the samples with health status, bacterial infections, and virus infections. Then, PAGE identified pairs of genes that exhibit significant differences. We quantified the dissimilarity between DGPs with a metric denoted $r_{dis}$, which captures the degree of difference between the expression profiles of DGPs. We then perform logarithmic operations on these values and subtract them to arrive at the difference between them. After determining the differences between individual DGPs using the PAGE algorithm, the next step is to validate the significance of these differences statistically. Subsequently, we employ Fisher's exact test to ascertain the p-value statistically, thereby identifying the 35 groups of DGPs that exhibit the most pronounced differential expression according to the following equation 1 and Table 2:

$$p = \frac{(a+b)!\cdot(c+d)!\cdot(a+c)!\cdot(b+d)!}{a!\cdot b!\cdot c!\cdot d!\cdot(a+b+c+d)!} \quad (1)$$

Table 1 | Frequency in Fisher's exact test

|  | Label<0.5 | Label>0.5 |
|---|---|---|
| $r_{dis}$ | a | b |
| $r_{dis}$ | c | d |

We procure 35 DGPs about distinct diseases, which are designated as the foundation for both training and validation datasets during model development.

**Model design.** In this model, we applied deep neural network with multiple self-attention layers. Self-attention mechanisms serve as an integral component in filtering pivotal information, thereby fostering the establishment of a comprehensive global dependency relationship, which in turn enhancing information acquisition. Multiple head self-attention mechanism is also the key for transformers. To emphasize the discrimination between DGPs and attain superior performance, a transformer layer was strategically incorporated into our model architecture. In single head self-attention mechanism, we use $X \in \mathbb{R}^{N*d_k}$ to represent our input infection data, where N is the total amount of data and $d_k$ to represent the feature dimension. Q, K, and V stand for Query, Key and Value, respectively. the output $Z$ can be obtained using equation 2. .

$$Z = softmax(\frac{Q \cdot K^T}{\sqrt{d_k}}) \cdot V \quad (2)$$

where $Q = W^Q \cdot X, K = W^K \cdot X, V = W^V \cdot X$ stand for Query, Key and Value, respectively. $W^Q$, $W^K$, and $W^K$ are learnable parameters.

The output of multiple self-attention mechanisms $Z_f$ is a splice of multiple $Z$ then mapped to the original space by a learnable linear transformation as follows with a weight matrix $W^f$.

$$Z_f = W^f \cdot \bigcup_{i=0}^{n} Z_i \quad (3)$$

Our framework comprises a meticulously designed teacher model, accompanied by two distinct student model structures. The teacher model comprises two transformer layers and a Multi-Layer Perceptron (MLP) component. The formula for updating the weights of the MLP is as follows:

$$w_{ji} \leftarrow w_{ji} + \Delta w_{ji} = w_{ji} + \eta \delta_i x_{ji} \quad (4)$$

where $w$ is the learnable weight, $x$ is input, $\eta$ is learning rate, and $\delta$ is local gradient.

Specifically, the first transformer layer incorporated a multi-head attention mechanism with 5 attention heads, the dropout rate is 0.1, and utilized the Gaussian Error Linear Unit (GELU)[48] activation function as follows:

$$Gelu(x) = 0.5 \cdot x \cdot (1 + \tanh(\sqrt{\tfrac{2}{\pi}}(x + 0.044715x^3))) \quad (5)$$

We completed the layer specification before the attention and feedforward operations and set up 4 encoder layers in the encoder. The second transformer layer, mirrored the structure of the first but with tailored configurations. It uses two attention heads in its multi-head attention module and two encoder layers in the encoder. Notably, it adopted the Rectified Linear Unit (ReLU)[49] activation function, catering to potential non-linearity requirements as following:

$$\text{ReLU}(x) = \max(0, x) \quad (6)$$

Additionally, a fully connected layer is interposed between the two transformer layers, aiming to augment the representational capabilities of the model by extending the feature space. Then we applied MLP with GELU activation within each linear layer and layer normalization is applied to facilitate learning stability.

We devised two distinct student models to address specific objectives and constraints. The first student model has one transformer layer which is the same as the teacher model's first transformer layer, including the multi-head attentional mechanism (5 heads), GELU activation, and layer normalization. It also has an MLP part. The second student model adopts a more streamlined structure, only comprising solely linear layers with GELU activation. The rationale behind designing these two student models is twofold: firstly, to validate the generality and adaptability of our Teacher-student structure across a range of datasets and tasks. Secondly, to accommodate the practical challenges posed by limited data availability, particularly in the context of streptococcal infection, where light models can have better performance. In conclusion, we designed the second student model for streptococcal infection data and used the first student model for other data.

**Training TSGPS.** Throughout the entire training process, we consistently utilize AdamW as the optimizer, owing to its effectiveness in handling large-scale and complex optimization tasks while incorporating weight decay for regularization. Initially, we train the teacher model on pan-infection data excluding specific conditions such as sepsis, HIV, RSV, etc. Subsequently, we use data of staphylococcal infection, streptococcal infection, HIV, RSV, and sepsis to train student models. During the training of the student models, which is tasked with predicting a distinct set of diseases, we present the teacher model with data identical to the student. The soft targets are obtained by applying a softmax function with a temperature parameter $\tau$ to smooth the distribution predicted by the teacher model. $\tau$ is set to 5 in the implementation process. To facilitate knowledge transfer from the teacher to the student, we construct a distillation loss function, denoted as $\mathcal{L}_{distill}$, which quantifies the discrepancy between the teacher's prediction $P^T$ and the student's predicted soft prediction $P^S$ as the following function:

$$q_i = \frac{\exp(P_i^T/\tau)}{\sum_j \exp(P_j^T/\tau)} \quad (7)$$

$$p_i = \log \frac{\exp(P_i^S/\tau)}{\sum_j \exp(P_j^S/\tau)} \quad (8)$$

$$\mathcal{L}_{distill} = \frac{\sum_i q_i (q_i - p_i)}{n \cdot \tau^2} \quad (9)$$

Where $n$ is the number of $p_i$.

Additionally, we use the cross-entropy loss $\mathcal{L}_{ce}$ to supervise the training of the student model based on the data-specific annotations $Y$ to preserve the specificity of the student model:

$$\mathcal{L}_{ce} = -\sum_i^n Y_i \cdot \log P_i^S \quad (10)$$

The total loss $\mathcal{L}_{Total}$ for the student network training is composed of the weighted combination of distillation loss $\mathcal{L}_{distill}$ and cross-entropy loss $\mathcal{L}_{ce}$:

$$\mathcal{L}_{Total} = w_{ditill} \cdot \mathcal{L}_{distill} + w_{ce} \cdot \mathcal{L}_{ce} \quad (11)$$

where $w_{ditill}$ and $w_{ce}$ are loss weight for distillation loss and cross-entropy loss, and set to 0.2 and 0.8 in practice, respectively.

TSGPS enables the student model to efficiently distill the knowledge in the teacher's predictions. Through this process, the student model achieves better performance even when dealing with limited data or specific disease prediction tasks.

To comprehensively evaluate the performance of these models after training, we employ a diverse set of evaluation metrics including AUC with the Receiver Operating Characteristic (ROC) curve, ACC, AUPRC, ROC, precision, and F1. The AUC, offering a holistic assessment of the classification model's overall efficacy, can be obtained through the False Positive Rate (FPR) and True Positive Rate (TPR) These metrics were calculated using functions as follows:

$$ACC = \frac{TP+TN}{TP+TN+FP+FN} \quad (12)$$

$$Recall = TPR = \frac{TP}{TP+FN} \quad (13)$$

$$FPR = \frac{FP}{TN+FP} \quad (14)$$

$$Precision = \frac{TP}{TP+FP} \quad (15)$$

$$F_1 = 2 \cdot \frac{Precision \cdot Recall}{Precision + Recall} \quad (16)$$

where TP is the positive example that was correctly predicted, TN is the correctly predicted negative example, FP is the positive example of the wrong prediction, and FN is the wrongly predicted negative example.

**Model deployment.** Our structure aimed at achieving an optimal balance between model quality enhancement and extreme parameter compression. To validate the ease of deployment of the student model, we conducted a thorough analyze their parameter count. Our analysis revealed that the teacher model boasts 18,142,949 parameters, the student model with transformer has 8,178,842 parameters (compression ratio 54.9%), and the student model after extreme compression has 797,925 parameters (compression ratio 95.6%). Notably, despite this substantial parameter reduction, the student model maintains a commendable performance in classification tasks. TSGPS has good performance in both normal and extreme compression and it makes real-world deployment possibilities.

**Data availability**
The data underlying this article are available in the GEO database, at https://www.ncbi.nlm.nih.gov/geo/. Data used for training and testing are available in Fig.2b.

**Acknowledgements**
This research was supported by National Natural Science Foundation of China (32370711, 62473191), Shenzhen Medical Research Fund (A2303033), High level of special funds from Southern University of Science and Technology, Shenzhen, China (G03034K003), Shenzhen Science and Technology Program (JCYJ20220530152409020, 20231115141459001), and the Shenzhen People's Hospital Clinical Scientist Foundation (SYWGSJCYJ202401).

**Author contributions**
L.C. conceived and designed the algorithms and experiments. H.W., L.Z., J.X., and Q.C. performed the analysis. X.Z., N.J., and C.Z. preprocessed the data and provided suggestions on improving the analysis. L.Z., and L.C. drafted the manuscript. X.Z., N.J, Q.C, and J.W. revised the manuscript. J.W. and L.C. supervised the study. All authors read and approved the final manuscript.